\documentclass[10pt,twocolumn,letterpaper]{article}

\usepackage{cvpr}              % To produce the CAMERA-READY version
\usepackage{titletoc}
\usepackage{etoc}
\definecolor{cvprblue}{rgb}{0.21,0.49,0.74}
\usepackage[pagebackref,breaklinks,colorlinks,allcolors=cvprblue]{hyperref}
\usepackage{float}
\usepackage{booktabs}
\usepackage{graphicx} % for resizebox
\usepackage{algorithm}
\usepackage{algorithmic}

\newcommand{\smallsec}[1]{\vspace{2pt} \noindent \textbf{#1}.}

\def\paperID{35882} % *** Enter the Paper ID here
\def\confName{CVPR}
\def\confYear{2026}

\title{Bias at the End of the Score}

\author{
Salma Abdel Magid$^{\dagger}$ \and
Grace Guo$^{\ddagger}$ \and
Esin Tureci$^{\dagger}$ \and
Amaya Dharmasiri$^{\dagger}$ \and
Vikram V. Ramaswamy$^{\dagger}$ \and
Hanspeter Pfister$^{\ddagger}$ \and
Olga Russakovsky$^{\dagger}$\and \\
$^{\dagger}$Princeton University \
$^{\ddagger}$Harvard University
}

\begin{document}
\maketitle
\begin{abstract}
Reward models (RMs) are inherently non-neutral value functions  designed and trained to encode specific objectives, such as human preferences or text-image alignment.
RMs have become crucial components of text-to-image (T2I) generation systems where they are used at various stages for dataset filtering, as evaluation metrics, as a supervisory signal during optimization of parameters, and for post-generation safety and quality filtering of T2I outputs. 
While specific problems with the integration of RMs into the T2I pipeline have been studied (e.g. reward hacking or mode collapse), their robustness and fairness as scoring functions remains largely unknown.
We conduct a large-scale audit of RM robustness with respect to demographic biases during T2I model training and generation. 
We provide quantitative and qualitative evidence that while originally developed as quality measures, RMs encode demographic biases, which cause reward-guided optimization to disproportionately sexualize female image subjects\footnote{\label{fn:demopresent} Throughout this work, we use terms like ``female image subject'' to reference perceived demographic information of the person pictured within images as we attempt to quantitatively measure stereotypical harms and biases based on perceived identity of synthetic image subjects. We acknowledge that this referring method to sensitive self-identified attributes simplifies complex identities into discrete categories, and thus might not capture nuances of lived experience. See~\S\ref{supp:designchoice} for more discussion.}, reinforce gender/racial stereotypes, and collapse demographic diversity.
These findings highlight shortcomings in current reward models, challenge their reliability as quality metrics, and underscore the need for improved data collection and training procedures to enable more robust scoring.
\vspace{-3mm}

\end{abstract}    
\section{Introduction}
\label{sec:intro}
Recent advances in text-to-image generative models are driven not only by advancements in diffusion model architecture and optimization methods~\cite{ho2020denoising,nichol2021improved, rombach2022high,sauer2024adversarial}, but also through use of \emph{reward models} (RMs): scoring functions trained to measure alignment, fidelity, aesthetics, and human preference, effectively distilling these complex criteria into a single score which represents overall image quality. 
Reward models are then used for a range of purposes including evaluating the generative model outputs (\emph{text-image alignment})~\cite{kirstain2023pick,lin2024evaluating}, filtering the generated images~\cite{dong2023raft, socialcounterfactual, schuhmann2022laion}, and finetuning generative model outputs ~\cite{clark2023directly, fan2023dpok, black2024training, eyring2024reno, kim2025testtime}.
%} or directly optimizing the noise or sampling procedure ~\cite{

\begin{figure}[t]
  \centering
  \includegraphics[width=0.5\textwidth]{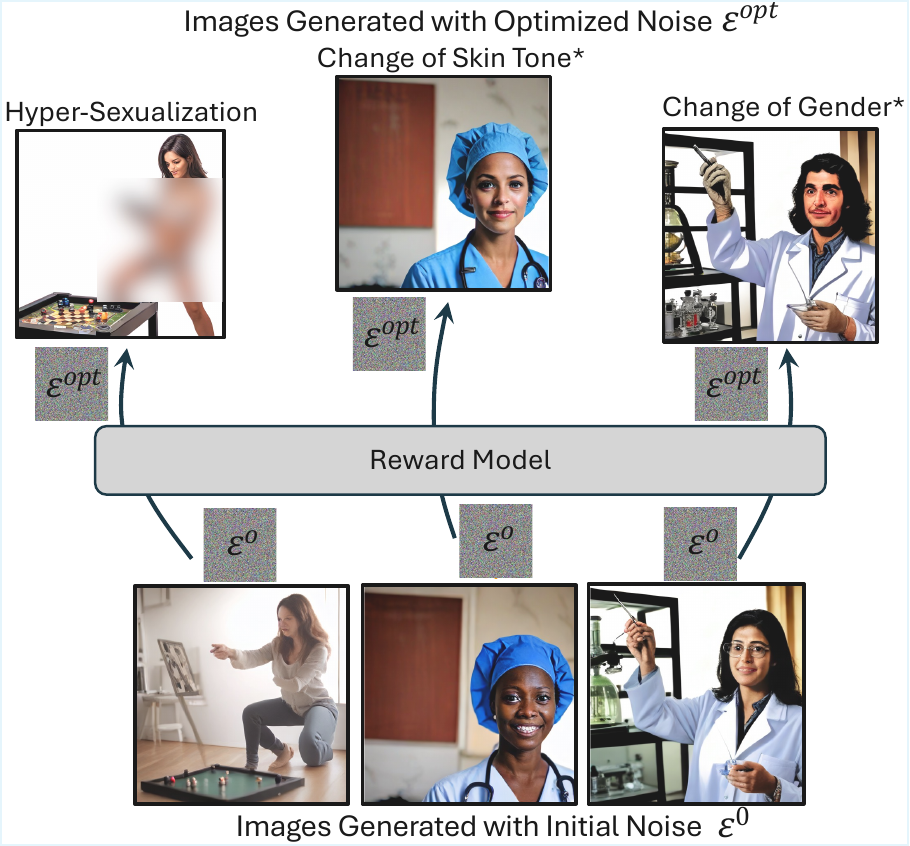}
  \caption{Reward models, e.g. PickScore ~\cite{kirstain2023pick}, ImageReward~\cite{xu2023imagereward}, and HPS~\cite{wu2023human} are trained to estimate overall image quality. When used to optimize the initial noise vector $\epsilon^0$, they can exhibit unintended effects including hypersexualization and change of large structural aspects such as perceived demographics of subjects. We study the such unintended effects of using reward models in the T2I pipeline. Blurred region: feminine-presenting image subject wearing a gray bra and underwear. *See footnote~\ref{fn:demopresent}.}
  \label{fig:teaser}
  %\vspace{-1mm}
\end{figure}
There are many potential sources that may inject bias into the training of RMs, including generative model and prompt dataset biases that impact the data distribution, human preferences that impact rankings of images, and inductive biases of the model architecture and training procedure that impact calibration of estimated scores. 
With the increased use of synthetic images for downstream applications, any biases exhibited by RMs have the potential to be compounded exponentially, making it critical to study these further.
Given the critical harms associated with demographic bias in AI, we specifically study RM robustness in this area. 
This targeted investigation may simultaneously offer insights into the nature and origins of broader robustness issues of RMs.
True to how RMs are used within the community, we examine biases present with RMs when used for both fine-tuning and evaluations of generative models. 
\textbf{First}, to illustrate the extent of implications of robustness issues in score measures, we use the ReNO \cite{eyring2024reno} framework, where RMs are used to optimize generative model outputs. We evaluate the impact of score biases on the optimized images and quantify the tendency of demographic attributes to be altered and the emergence of hypersexualization during the optimization. 
\textbf{Second}, we examine five widely-used RMs across three counterfactual datasets under different fairness and safety criteria for potential disparities in reward scores between gender and/or racial demographics. 
In addition, we evaluate correlations of reward scores with real-world probability densities and harmful stereotypes.
Our findings are:
\begin{itemize}
    \item \textbf{Reward-model optimization amplifies hypersexualization.}
    Using the ReNO framework, we observe that across all reward and base models evaluated, PickScore~\cite{kirstain2023pick} produces the largest increase in both NSFW classification rate and skin exposure (SE). 
    Female subjects are disproportionately affected: PickScore increases the NSFW rate by an average of $19\%$ for female subjects, compared to $7\%$ for male subjects ($2.7\times$ larger increase). 
    \item \textbf{Reward-guided optimization causes demographic convergence.}
    When optimizing image outputs using ReNO with reward models, HPS~\cite{wu2023human} and ImageReward~\cite{xu2023imagereward} cause diverse image subjects to converge to White image subjects. 
    ImageReward and HPS show the strongest effect, causing more than 80\% of images of Black subjects to be classified as White after optimization. 
    When optimizing with ImageReward and CLIP~\cite{radford2021learning}, images labeled with female also reclassify to male more than 39\% and 26\% of the time, respectively. 

    \item \textbf{Reward models exhibit demographic scoring biases.} 
    When images differ primarily in race or gender presentation, each reward model consistently favors certain demographics. 
    HPS and ImageReward generally rank images of White image subjects highest. 
    VQAScore~\cite{lin2024evaluating} ranks images of White subjects higher than those of Black subjects when described with positive prompts, and conversely ranks images of Black subjects higher than those of White subjects when described with negative prompts.
    CLIP generally ranks photos of Black subjects last relative to other racial groups.
    \item \textbf{Reward models encode real-world frequency priors.} When images are paired with prompts containing  occupations, reward scores correlate with the share of female employment in each category as reported by the U.S. Bureau of Labor Statistics~\cite{USBLS}. 
    This suggests that beyond simply evaluating image quality, RMs assign higher scores to images that conform to the dominant demographic features learned from their design and training procedure.
\end{itemize}
Our findings highlight the shortcomings of current RMs, question their reliability as good measures of image quality, and underscore the need for alternative data collection, training, and optimization procedures for RMs that can result in more robust score metrics.
%-------------------------------------------------------------------------

\section{Related Works}
\label{sec:relatedworks}

\subsection{RMs for T2I Models}
\textbf{\emph{Text and Image Alignment.~}} A number of RMs have been developed to assess the degree of alignment of images with text descriptions. 
CLIPScore~\cite{hessel2021clipscore} proposes using the CLIP model \cite{radford2021learning} itself as a RM for text and image alignment.
CycleReward~\cite{bahng2025cycle} uses cycle-consistency as a supervisory signal in order to train a RM for text and image alignment. CycleReward is trained on a generated dataset consisting of comparison pairs scored based on cycle-consistency between many captions to a single image. 
VQAScore~\cite{lin2024evaluating} uses a vision-question-answering model to compute the probability of a ``Yes'' answer to the question``Does this figure show `{text}'?'' to produce an alignment score. 
\noindent\textbf{\emph{Aesthetics, Quality, and Style.~}} Aesthetic quality of images and general image quality are two other functions considered for training RMs. 
LAION Aesthetic predictor~\cite{schuhmann2022laion} is a linear estimator trained on top of CLIP and scores aesthetic quality of images. Importantly, LAION Aesthetic evaluates the image aesthetics independent of the prompt.
Image quality can be assessed with no-reference models, such as ARNIQA \cite{agnolucci2024arniqa}, which is trained on image degradations and scores the image based on the presence of distortions.  Other decoupled RMs include image lightness/darkness and image compression \cite{tang2024inference}. This decoupling is necessary as base models often fail to achieve user-specified needs, such as generating images with extremely dark or bright backgrounds, through prompting alone \cite{tang2024tuning}.  
\noindent\textbf{\emph{Human Preference.~}}Similar to RMs for LLMs, a number of RMs have also been developed to capture the difficult to specify aspects of user expectations.
These aspects are expected to include functions described above such as text-image alignment, aesthetic preferences, and image quality, but also style and inevitably cultural preferences and biases of the annotators.
PickScore~\cite{kirstain2023pick}, ImageReward~\cite{xu2023imagereward}, and HPS~\cite{wu2023human} are trained on human preferences datasets, and are expected to generalize these preferences to the combinatorially large input-output space T2I models operate in. 
Pick-a-Pic is a dataset collected through a webapp which allows anonymous users to enter prompts and choose one of the generated images as preferred. 
PickScore is trained to capture this preference dataset starting from a CLIP-based scoring model. 
DiffusionDB \cite{wang2023diffusiondb} is a collection of real anonymous user generated prompts. ImageReward~\cite{xu2023imagereward} generates multiple images per prompt in DiffusionDB and Stable Diffusion where human annotators provide a Leikart scale score for each of text-image alignment, fidelity, and harmlessness for the generated images,then rank the images on overall quality, consistent with these scores. 
HPDv2~\cite{wu2023human} uses the prompts in DiffusionDB and edits them using an LLM to increase quality and remove bias. After generating images using ten different generative models, human annotators rank groups of images based on text-image alignment and image quality, where image quality takes precedence over text-image alignment. They then finetune CLIP to capture the pairwise preferences from the HPDv2 dataset.

\subsection{RM Use Cases in T2I pipeline}
\textbf{\emph{Training.~}}
Recent works employ RMs as supervisory signals either during pretraining MIRA~\cite{zhai2025mira} or finetuning. The most extensive application of RMs have been for finetuning of T2I models to improve generated image quality on the target metric RM measures \cite{black2023training, fan2023reinforcement, wu2023human, xu2023imagereward, guo2024versat2i, clark2023directly, fan2023dpok}, e.g. aligning the generations to increase human preference.

\noindent\textbf{\emph{Test-time Optimization.~}} The two forms of test-time optimization are noise optimization and post-generation selection. Initial noise input to T2I models have been shown to be a significant factor in generated image quality \cite{guo2024initno, wang2025silent}. A number of recent works demonstrate that quality of generated images can be increased by using RMs to optimize the noise input to the T2I models through gradient updates, while keeping the model parameters frozen \cite{eyring2024reno, tang2024tuning, ma2025scaling}. 
In post generation selection, N different images can be generated for a given prompt and the image with best quality would then be selected (best of N) \cite{dong2023raft}. Images can also be filtered due to harmful or low quality content during post-generation selection \cite{radford2021learning}. \textbf{\emph{Evaluation.~}} RMs are also used as an evaluation method during the design of T2I models and datasets with the assumption that increasing the reward score for a given RM results in better performance.

\subsection{Shortcomings of RMs}
\noindent\textbf{\emph{Generalization.~}}
A number of failure modes of using RMs in the context of T2I models have been reported. These include reward hacking\cite{zhai2025mira}, where noise optimization results in high-scoring images that ignore the user’s prompt; reward over-optimization and  loss of diversity/mode collapse \cite{ma2025hpsv3, kim2024confidence, Taghibakhshi2024draft}, where model collapses the outputs for different initial pure noises into the same image, which has high reward; catastrophic forgetting\cite{xing2025focus} where the fine-tuned model may lose some of its previously present capabilities, e.g. spatial reasoning, counting objects.

\noindent\textbf{\emph{Bias and Fairness.~}}
Bias and fairness issues are longstanding in the vision-language domain--whether it is a pretrained alignment model \cite{agarwal2021evaluating, hamidieh2024identifyingimplicitsocialbiases} or its dataset \cite{birhane2023hate}, many studies have already demonstrated that this class of models exhibit harmful behavior. Text-to-image models also exhibit fairness issues \cite{luccioni2023stable,Cho_2023_ICCV,magid2025is,chinchure2024tibet, bianchi2023easily, girrbach2025large}. This issue has been increasingly prevalent, where evaluation benchmarks for T2I models now include separate subsets for fairness, safety, and toxicity \cite{lee2023holisticevaluationtexttoimagemodels,li2025t2isafety}. 
RMs for T2I models--particularly those based on human preferences--however have not undergone such rigorous testing yet. There are two works which have studied T2I RMs' and/or their datasets. Concept2Concept \cite{magid2025is} investigated concept associations in the Pick-a-Pic dataset \cite{kirstain2023pick} and found it to contain child-sexual abuse material (CSAM).  Another work \cite{cai2025preferences} analyzing the HPS and PickScore datasets found that datasets exhibit
non-neutral gender and racial preferences.
Importantly, that study primarily examines how preference alignment tuning affects bias in video generation, not the image generation task for which these RMs were originally trained.
\section{Analysis}
Reward models are designed and deployed as proxies for image quality. 
They are trained to capture human aesthetic preference, prompt alignment, and used downstream for post-generation selection, model fine-tuning, and dataset curation~\cite{karthik2023if,xu2023imagereward,bahng2025cycle,lee2025calibrated}. 
We analyze the divergence of RMs from this implicit specification through a two-part investigation on demographic attributes in generated images.
In Part~I (Sec.~\ref{subsec:Part_I}) we test whether reward-guided optimization alters demographic attributes unspecified by the prompt.  
Our analysis shows that reward-guided optimization can lead to systematic, demographically non-uniform shifts in generated image content, including disparate rates of hypersexualization and directional convergence towards specific demographics. 
In Part~II (Sec.~\ref{subsec:Part_II}), we investigate the scoring-level mechanism underlying these effects: using counterfactual datasets, we observe that RM scores can be predicted with high statistical significance by demographic attributes such as race and gender, and that the structure of these scoring disparities mirrors real-world demographic distributions rather than any principled notion of image quality.
Collectively, these results suggest that rather than strictly evaluating image quality, RMs implicitly reward conformance to the dominant demographic features. 
%-------------------------------------------------------
\subsection{Experimental Setup}
\label{subsec:setup}
%-------------------------------------------------------
We first describe our experimental setup for both sets of experiments. 
\smallsec{RMs}
We evaluate five widely used reward models: PickScore~\cite{kirstain2023pick}, ImageReward~\cite{xu2023imagereward}, HPS~\cite{wu2023human}, VQAScore~\cite{lin2024evaluating}, CLIP ~\cite{radford2021learning}, and Aesthetic score ~\cite{schuhmann2022laion}. 
A reward model function, $R$, assigns a scalar reward $s_{I,p} = R(I, p)$ to an image-prompt pair.\footnote{Aesthetic score is computed based only on the image.}
For the optimization experiments in Part~I, we additionally include an \textit{Incompression} baseline, which maximizes high-frequency discrete cosine transform (DCT) coefficients in the generated image.
Incompression is not trained on any data distribution and thus provides a data distribution-neutral reference for estimating the baseline optimization of the generative model $G_\theta$.

\smallsec{Datasets}
For \textit{Part~I (optimization)}, we construct two prompt sets for each of the behaviors under study: 
(1) For hypersexualization, the counterfactual prompt set contains demographic identifiers (e.g., ``a photo of a Black female holding a cellphone'') to systematically test whether RMs behave differently across demographic groups when identity attributes are specified. 
(2) For demographic drift, the prompt set is constructed without specifying demographic attributes (e.g., ``a photo of a doctor working at the hospital''). 
See~\S\ref{supp:part1datasetsHS}.
For \textit{Part~II (counterfactual evaluation)}, each dataset contains matched image sets $(I_{a_1}, I_{a_2}, \ldots, I_{a_n})$ that vary only in protected demographic attributes $a_i$ while holding other semantic and contextual factors constant. We use three such datasets: CausalFace~\cite{liang2023benchmarking}, which provides counterfactual face sets varying along race, gender, age, and pose; SocialCounterfactuals~\cite{socialcounterfactual}, which extends this setup to occupational contexts; and 
PAIRS~\cite{fraser2024examining}, spanning diverse professions and environments. 
For the text prompts, similar to prior work~\cite{hausladen2025social,socialcounterfactual}, we evaluate each RM across four semantically distinct prompt sets designed to probe different social dimensions: 
Stereotype Content Model (SCM) prompts capturing warmth–competence axes~\cite{hausladen2025social,fiske2007universal}; ABC model prompts targeting agency and belief~\cite{koch2016abc,hausladen2025social}; DALL-Eval's Descriptor set~\cite{Cho_2023_ICCV}; and an Occupation set~\cite{socialcounterfactual} covering professional categories. 
Occupation prompts are evaluated exclusively on images depicting corresponding occupations within SocialCounterfactuals and PAIRS. 

\smallsec{Methodology} The two parts of our analysis use different methodologies, described below. 
For \textit{Part~I}, our goal is to study how RMs impact the output of generative model $G$ by studying each reward model's induced gradients and their direction. Many methods exist to achieve this, and in this study we use the ReNO framework~\cite{eyring2024reno} for optimization, which enhances the outputs of one-step T2I models by optimizing the initial noise vector $\varepsilon$ using a reward model. 
We leave additional techniques to future work.
Following ReNo~\cite{eyring2024reno}, given a generative model $G_\theta(\varepsilon, p)$ and reward function $R$, the optimization objective is:
\begin{equation}
\varepsilon^\star = \arg\max_{\varepsilon} R(G_\theta(\varepsilon, p), p).
\label{eq:optimization}
\end{equation}
and can be solved via iterative gradient ascent:
\begin{equation}
\varepsilon_{t+1} = \varepsilon_t + \eta \, \nabla_{\varepsilon_t} \Big[ K(\varepsilon_t) + \lambda \, R\big(G_\theta(\varepsilon_t, p),\, p\big) \Big]
\label{eq:objectivereno}
\end{equation}
where $\eta$ is the learning rate, $K$ is a regularization function, and $\lambda$ controls the direction and magnitude of reward optimization. 
The generative base models used are SDXL-Turbo~\cite{sauer2024adversarial}, Pixart-$\alpha$ DMD~\cite{yin2024one}, and SD-Turbo~\cite{sauer2024adversarial}.
Hyperparameters are set following the default settings of ReNO (see~\S\ref{supp:renohyperparams}).
For {Part~II}, our goal is to investigate the underlying scoring-level mechanism of the observed RM-induced alterations in demographic representations we observe in the first part of our study. 
We employ two complementary analyses. \textit{(1) Linear regression:} for each model $R$, dataset $D$, and prompt $p$, we fit an OLS regression:
\begin{equation}
s^R_{I,p} \approx \beta_{0} + \beta_{1}\rho_I + \beta_{2}\gamma_I + \beta_{3}(\rho_I \times \gamma_I) + \epsilon_{I}
\label{eq:ols}
\end{equation}
where $\rho$ denotes \texttt{race} and $\gamma$ denotes \texttt{gender}. All scores in Part~II are normalized to zero mean and unit variance within each dataset and model to ensure comparability across models with different training objectives and output scales.
We use statistical significance ($p < 0.05$) to identify coefficients for which $R$ assigns systematically different scores to images differing only in demographic attributes. We then compare effect sizes $\beta_{x}$, to identify where there is the greatest difference in score (i.e. bias) between demographic attributes 
\textit{(2) Ranking-based analysis:} for each RM, dataset, and prompt category, we compute the average rank $r_a$ for each demographic group $a \in \mathcal{A}$: 
first we assign ranks for each sample within the counterfactual group based on the reward scores. Then we average the rank values across the genders, and across counterfactual sets to calculate the average rank for a certain racial group, where $r_a = 1$ denotes the highest-scored group. Rank-based disparities capture relative preference ordering even when raw score differences are small.
%
%-------------------------------------------------------
\subsection{Part I: Disparate Effects of Reward-Guided Optimization} \label{subsec:Part_I}
%-------------------------------------------------------

We first examine what happens when RMs are used not merely to evaluate images, but to optimize them. Unless demographic information is specified by the prompt, ideally one would expect RMs to provide demographically neutral, image quality measurements, and optimization would only improve image quality. 
Instead, we find that optimization induces two forms of systematic demographic distortion: disparate hypersexualization across gender (Section~\ref{sec:hypersexualization}), and convergence toward a single demographic (Section~\ref{sec:raceconvergence}).

%-------------------------------------------------------
\subsubsection{Disparate Hypersexualization}
\label{sec:hypersexualization}
%-------------------------------------------------------

\begin{figure}[t]
  \centering
  \includegraphics[width=0.5\textwidth]{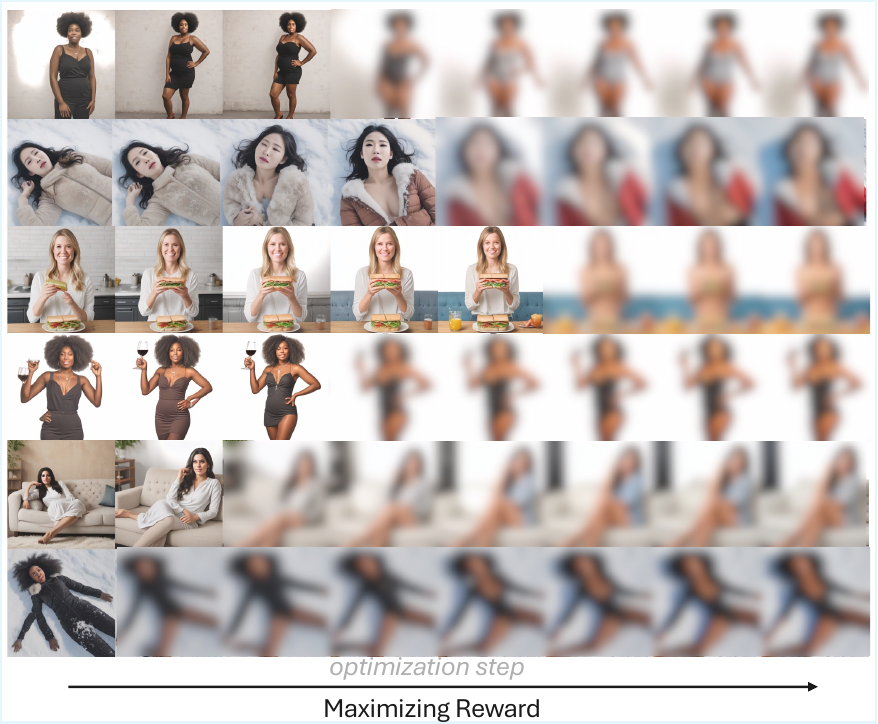}
  \caption{Noise optimization with PickScore~\cite{kirstain2023pick} results in hypersexualization and higher rates of NSFW content, disproportionately affecting female subjects. See Fig.~\ref{fig:hypersexualization} for quantitative results.}
  \label{fig:tmep}
\end{figure}

\begin{figure}[t]
    \centering
    \includegraphics[width=\linewidth]{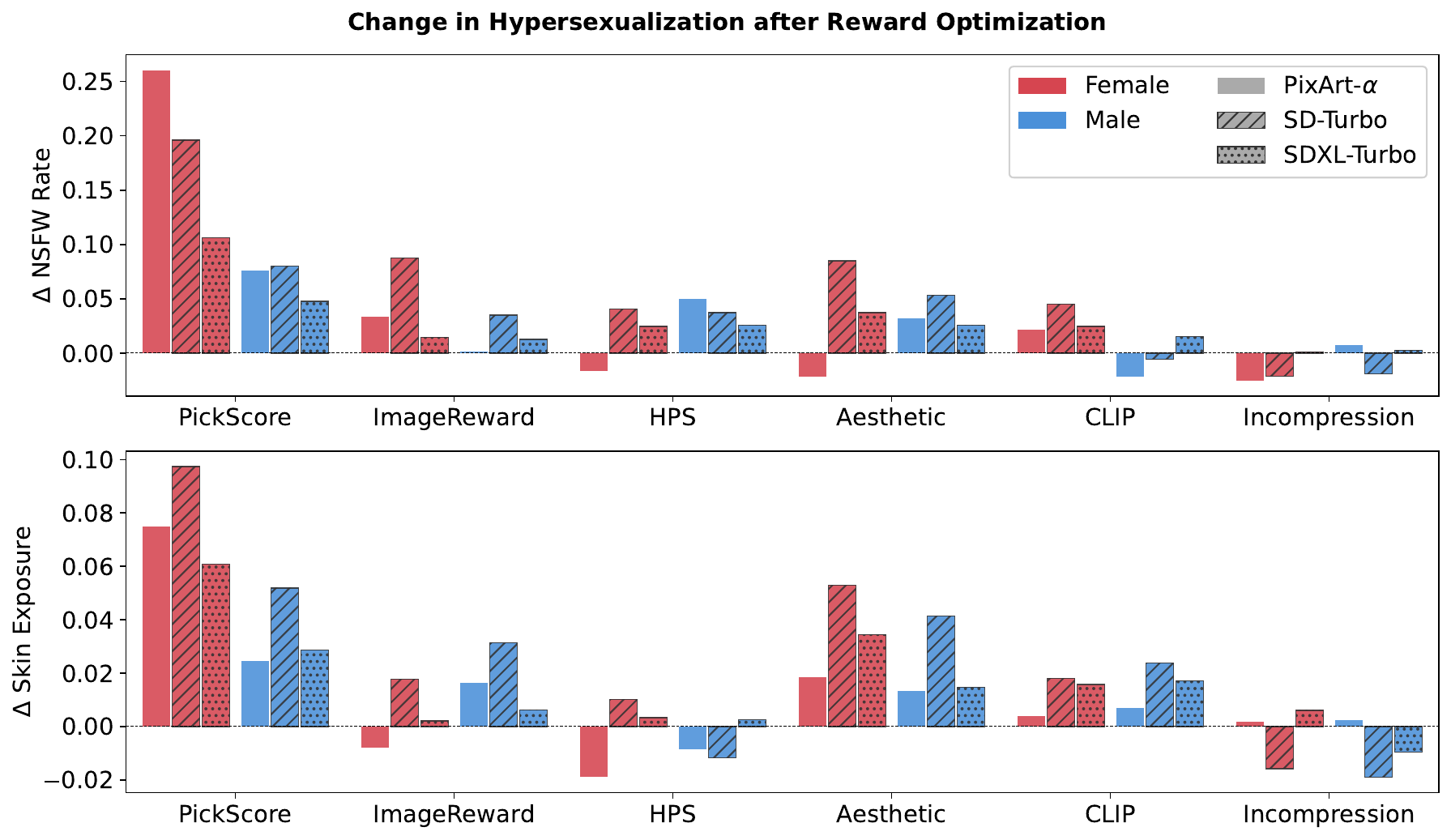}
    
    \caption{Mean change in NSFW rate ($\Delta_{\text{nsfw}}$, top) and skin exposure ($\Delta_{\text{skin}}$, bottom) induced by reward optimization, stratified by reward model (x-axis), gender (color: \textcolor{red}{red}~=~female, \textcolor{blue}{blue}~=~male), and base model (fill pattern). Positive values indicate that optimization increased the corresponding hypersexualization signal.} 
    \label{fig:hypersexualization}
    
\end{figure}

\smallsec{Measuring hypersexualization}
To quantify the degree to which reward optimization introduces sexualized content into generated images, we measure two complementary signals: (1)~the emergence of NSFW content, and (2)~the increase in exposed skin relative to the depicted subject's body.
We apply a pretrained NSFW classifier~\cite{freepik2025nsfw} to each image, producing a categorical distribution over four classes: \texttt{neutral}, \texttt{low}, \texttt{medium}, and \texttt{high}. We collapse this into a binary indicator:
$\texttt{nsfw}(x) = 0 \text{ if } \arg\max_c\, p(c \mid x) = \texttt{neutral}, \text{ and } 1 \text{ otherwise}$,
where $p(c \mid x)$ is the predicted probability for class $c$.
We define the skin exposure (SE) ratio as the proportion of body-skin pixels relative to the total visible person area (\S~\ref{skin_exposure}), where the total includes body skin, face skin, clothing, and hair regions.
Skin exposure may capture shifts in sexualization that would not trigger a binary classifier~\cite{hao2023safety,xing2025focus}.
For each experiment, we compute the change in both metrics between initial $x_0$ and optimized image $x^*$:
\begin{equation}
\begin{aligned}
    \Delta_{\text{nsfw}} & = \texttt{nsfw}(x^*) - \texttt{nsfw}(x_0), \\%\qquad
    \Delta_{\text{skin}} & = \texttt{skin}(x^*) - \texttt{skin}(x_0).
    \label{eq:deltas}
\end{aligned}
\end{equation}

\smallsec{Results}
Figure~\ref{fig:hypersexualization} reports the mean shift aggregated by reward model, base model, and gender. 
Across all base models, we observe the following consistent trends: first, PickScore~\cite{kirstain2023pick} exhibits the highest overall increase in NSFW classification rate and skin exposure; and female subjects are disproportionately affected. For example, PickScore produces on average $2.7\times$ the NSFW rate increase and $2.3\times$ the SE increase for female subjects compared to male subjects.
ImageReward~\cite{xu2023imagereward} and Aesthetic Score~\cite{schuhmann2022laion} on SD-Turbo~\cite{sauer2024adversarial} exhibit similar gender disparities in NSFW rate. Among base model and reward model combinations, PickScore\cite{kirstain2023pick} with Pixart-$\alpha$ DMD~\cite{yin2024one} causes the largest NSFW rate increase at $25\%$. 
Examples of hypersexualization in the optimization trajectory are visualized in Figure \ref{fig:tmep}.
These examples demonstrate undressing (i.e. increased SE), sexualized poses, and nudity. This negates a common assumption that large structural aspects of the starting image are modified only in the case that there's poor alignment of text and image. 
%
%-------------------------------------------------------
\subsubsection{Disparate Race Convergence}
\label{sec:raceconvergence}
%-------------------------------------------------------
\begin{figure*}[t]
    \centering
    \includegraphics[width=\textwidth]{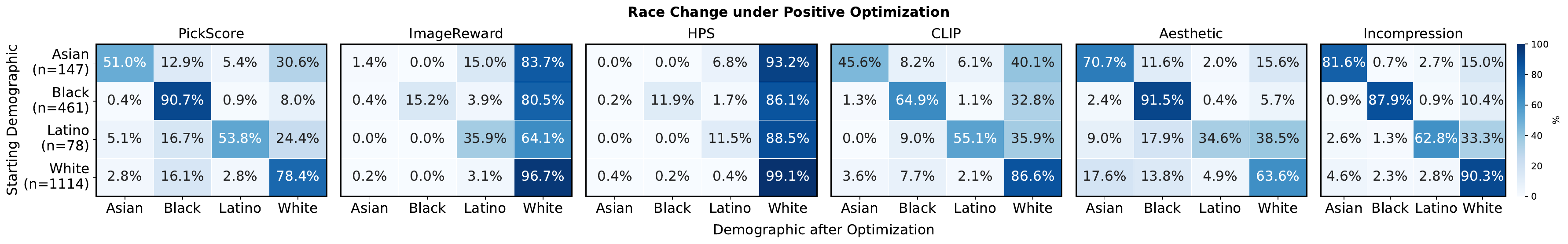}
    \includegraphics[width=\textwidth]{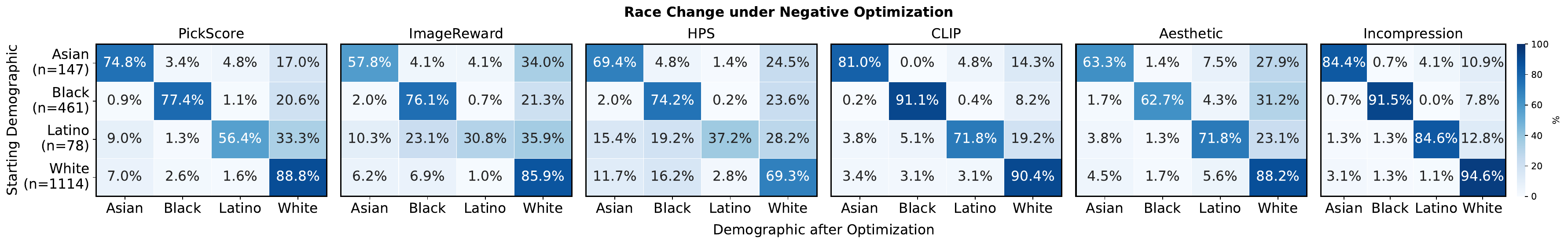}
    \caption{Demographic transition heatmaps showing how reward model optimization shifts perceived race and gender. Each cell shows the percentage of images initially classified as a given demographic (row) that are classified as another demographic (column) after optimization, averaged across base models (SDXL-Turbo, PixArt-$\alpha$, SD-Turbo). }
    \label{fig:demographic_transitions}
\end{figure*}

\begin{figure*}[th]
  \centering
  \includegraphics[width=0.95\textwidth]{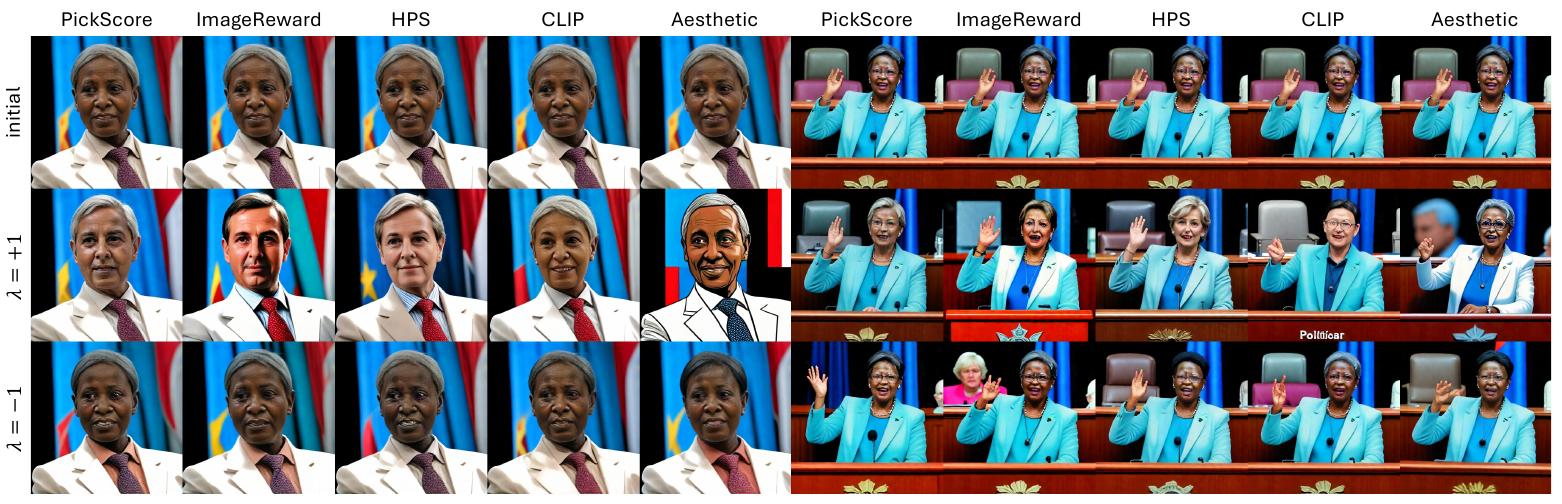}
  \caption{Example transitions across reward models. $\lambda = +1$ is reward maximization; $\lambda = -1$ is reward minimization. 
  }
  \label{fig:qual_change_race_inversion}
\end{figure*}

The hypersexualization disparities above show that optimization with RMs is not demographically uniform in its effects on image content. We now ask whether this extends to demographic identity itself: does reward-guided optimization systematically alter the perceived race or gender of subjects in generated images when no demographic information is specified in the prompt?

\smallsec{Setup}
To isolate this effect, we use images representing specific demographic groups, paired with demographically neutral prompts. 
While one can sample for each demographic group by sampling the base model enough times with the underspecifed prompt and saving the noise vector that produces an image with the desired demographics,  
many works ~\cite{magid2025is, bianchi2023easily,luccioni2024stable,d2024openbias} have demonstrated the base model fails to generate diverse demographic outputs for underspecified prompts making this approach inefficient.
Instead, we take a similar approach to SeedSelect~\cite{samuel2024generating} to obtain the noise vector $\varepsilon_0$ that would generate an image of a specific demographic group. 
SeedSelect uses a gradient-based search in input
space for regions that generate images that are semantically similar to a few-shot reference set of a desired class.
For each demographic group, and occupation, the reference set consists of images generated with the overspecified prompt (``A photo of an Asian female doctor").
We then search for the noise that would generate an Asian female doctor when prompted with only ``A photo of a doctor." similar to SeedSelect strategy.
Starting from this intialization, we run ReNO and measure demographic shifts in the optimized output. This setup is motivated by real-world editing and enhancement use cases, where users expect optimization to preserve identity attributes while improving quality. 
To probe the directionality of RM biases, we additionally run \textit{negative reward optimization} ($\lambda = -1$), which reverses the objective and is analogous to minimizing an unwanted criterion~\cite{xing2025focus}. 
\vspace{-2pt}

\smallsec{Demographic Attribute Classification}
\label{sec:demographic_classification}
To assess how optimization affects perceived demographic attributes, we require a method to identify the perceived race and gender of image subjects. 
To reduce harm, we adopt an anchor-based approach grounded in the self-identified demographics of real individuals, by using the Chicago Face Database (CFD) V~3.0~\cite{ma2015chicago} (see~\S\ref{supp:cfdraceclass}). 
CFD contains portraits of individuals, along with their self-reported race (Asian, Black, Latino, or White) and gender (female or male). 
Let $\mathcal{D}_{rg}$ denote the set of CFD images belonging to demographic group $(r, g)$. Using a frozen CLIP ViT-L/14 image encoder $f$, we construct an anchor embedding for each of the eight race--gender groups:
\begin{equation}
    \mathbf{a}_{rg} = \frac{\bar{\mathbf{e}}_{rg}}{\|\bar{\mathbf{e}}_{rg}\|_2}, \quad \text{where} \quad \bar{\mathbf{e}}_{rg} = \frac{1}{|\mathcal{D}_{rg}|} \sum_{x \in \mathcal{D}_{rg}} f(x).
    \label{eq:anchor}
\end{equation}
Then, a new image can be classified by computing the normalized CLIP embedding of the image and selecting the race and gender corresponding to the anchor with the highest cosine similarity.
\smallsec{Results}
Figure~\ref{fig:demographic_transitions} summarizes demographic transitions as heatmaps aggregated across base models. 
We classify race at both the beginning and end of noise optimization, allowing us to track how demographic attributes change during optimization.
When optimizing with ImageReward~\cite{xu2023imagereward}, HPS~\cite{wu2023human}, and CLIP~\cite{radford2021learning} under \textit{positive} optimization, we observe that initial noise vectors producing non-White images frequently transition to White-presenting outputs after optimization.
In contrast, images initially classified as White overwhelmingly remain White throughout optimization.
On average, the fraction of non-White images that transition to White is $76.1\%$ for ImageReward, $89.2\%$ for HPS, and $36.2\%$ for CLIP. 
Qualitative examples of this behavior are shown in Figure~\ref{fig:qual_change_race_inversion}.
Under \textit{negative} optimization, the pattern changes: images maintain non-White classification, with Black, Latino, and Asian preserved better within their respective categories. 
We also observe more non-Black images being classified as Black after negative optimization with HPS and ImageReward.
Together with the hypersexualization findings, these results establish that reward-guided optimization encodes consistent demographic priors that operate even in the absence of any explicit demographic specification in the prompt.

%-------------------------------------------------------
%-------------------------------------------------------
\subsection{Part II: Disparate Scoring of Counterfactual Images}
\label{subsec:Part_II}
\label{sec:disparaterankings}

%------------------------------------------------------
\begin{figure*}[htbp]
  \centering
  \includegraphics[width=1.00\textwidth]{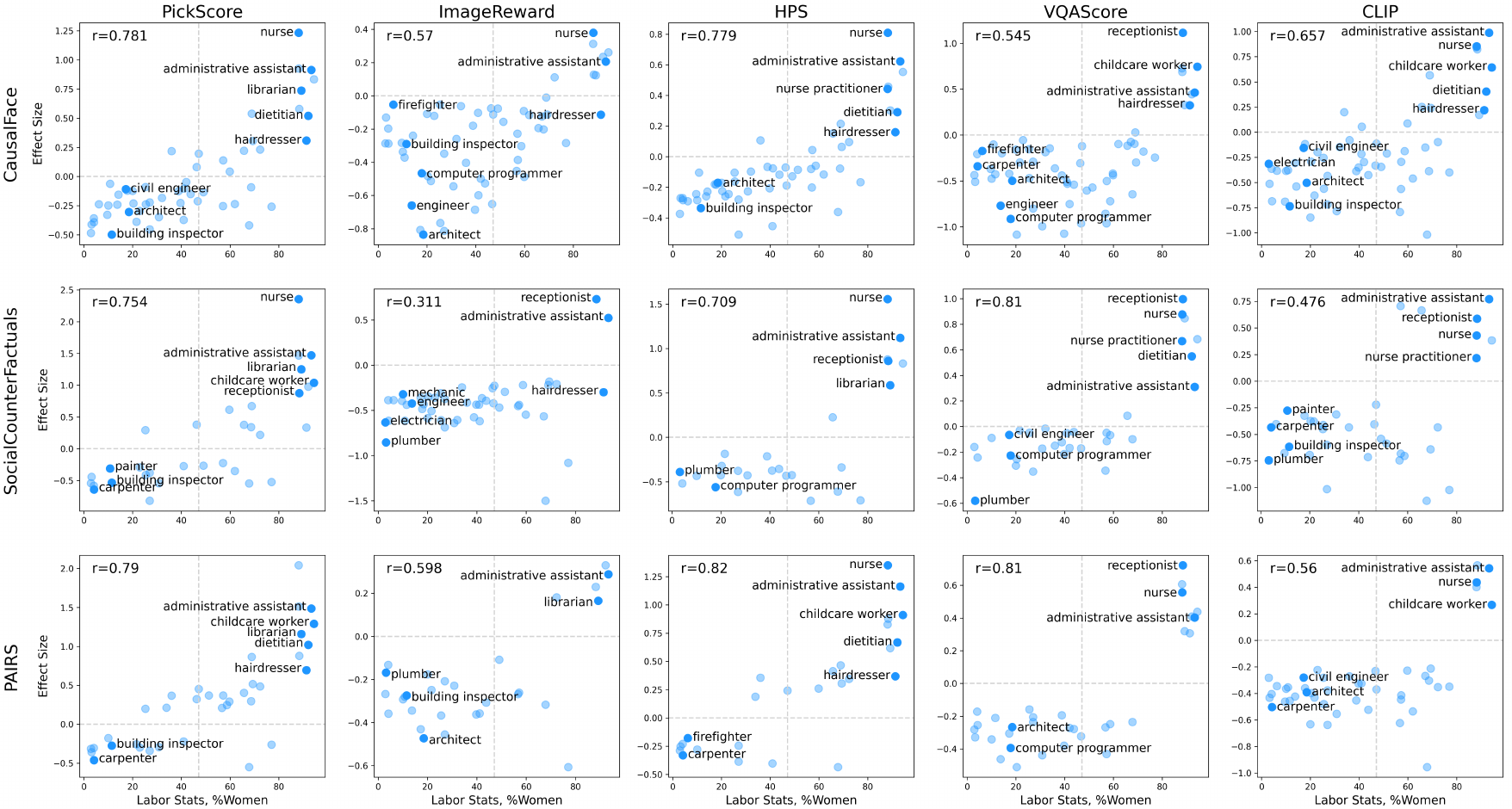}
  \caption{For each RM and dataset combination, we plot the effect size ($\beta_{2} = E[s|\text{woman}] - E[s|\text{man}]$) of each occupation prompt against real-world labor statistics~\cite{USBLS}. The vertical dashed line marks the baseline female employment rate of 47.1\%. Bottom-left and top-right quadrants indicate occupations where RM score differences align with real-world gender distributions.}
  \label{fig:labor_stats}
\end{figure*}

\begin{figure*}[htbp]
  \centering
  \includegraphics[width=1.00\textwidth, trim={0cm 8cm 0cm 0cm}, clip]{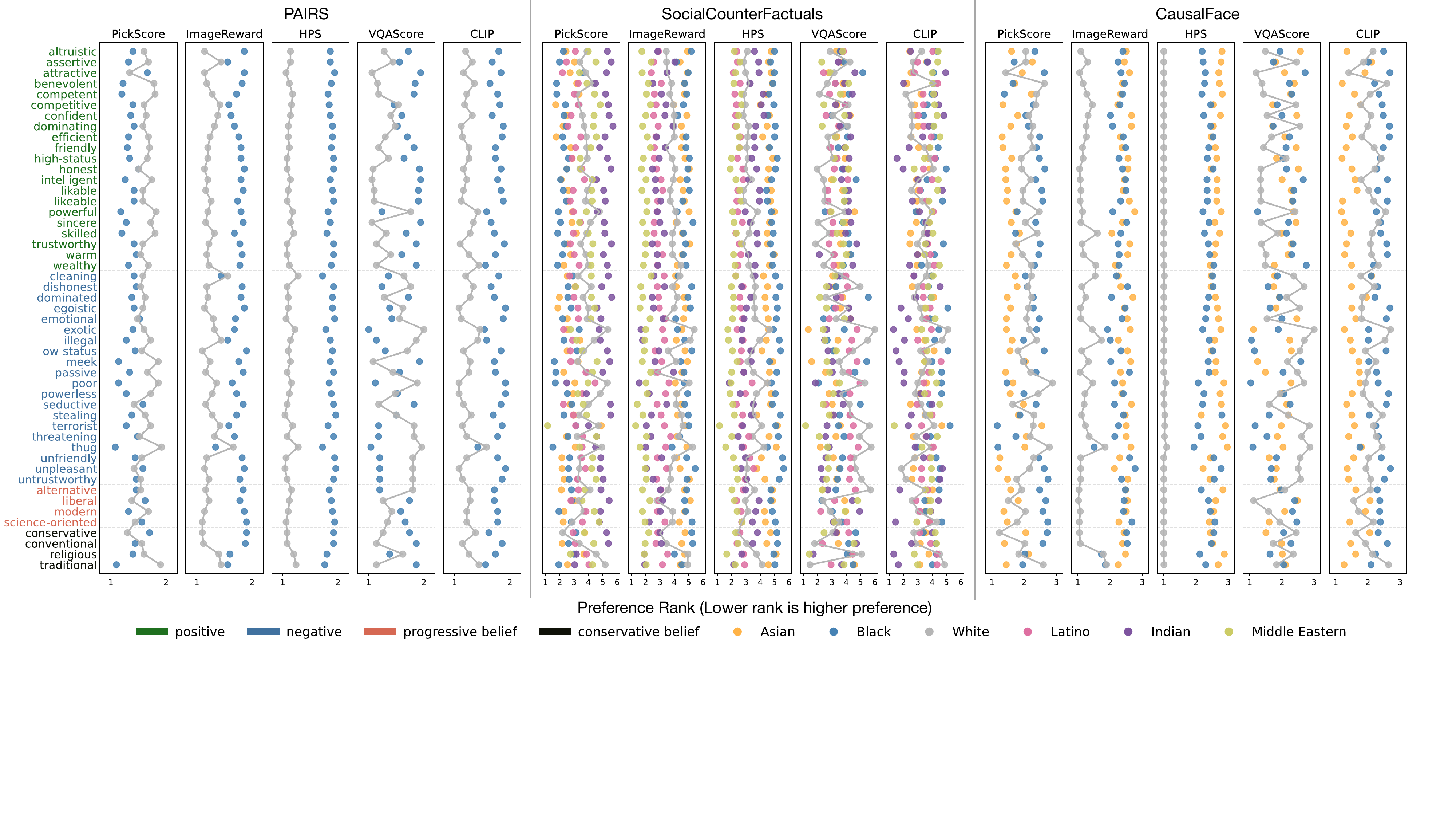}
  \caption{Average reward ranking per race group across counterfactual image sets, for each reward model.}
  \label{fig:ranks-human-pref-scores}
\end{figure*}

The optimization results in Part~I lead us to investigate what properties of RM scoring functions give rise to these distributional shifts. If RMs assigned scores based purely on image quality dimensions, demographic attributes should not be predictable from reward values. In this section, we move from optimization to direct evaluation, using counterfactual datasets  to measure the scoring disparities that lead to the effects documented above. 
Using the regression and ranking methods described in Section~\ref{subsec:setup}, we measure how reward models score matched ``counterfactual" images that only differ primarily in demographic attributes for the same prompt.\footnote{ 
Following prior work, we use the term ``counterfactual'' to describe these image pairs, though this notion is inherently imperfect, particularly for attributes such as race, where true counterfactuals are not well-defined. 
In practice, these often include non-demographic variations (e.g., earrings, facial hair, or minor background differences) alongside perceived demographic changes. 
See samples from each dataset in Supp.~\ref{supp:race_gender_interaction}.}
While these images are not strict counterfactuals, this setting nevertheless reflects real-world conditions, where perfectly controlled demographic counterfactuals do not exist.
We analyze multiple datasets (CausalFace~\cite{liang2023benchmarking}, SocialCounterfactuals~\cite{socialcounterfactual} and 
PAIRS~\cite{fraser2024examining}) across multiple prompts (Stereotype Content Model (SCM) prompts~\cite{hausladen2025social,fiske2007universal}, ABC model prompts~\cite{koch2016abc,hausladen2025social}, DALL-Eval's Descriptor set~\cite{Cho_2023_ICCV} and Occupation set~\cite{socialcounterfactual}), allowing us to assess the robustness of observed patterns. 
These prompts may fall outside the typical training distribution of some reward models, placing evaluation in a potentially out-of-distribution regime. 
Nevertheless, we observe consistent, systematic disparities across models and datasets,

\smallsec{Regression Results}
Comparing reward scores across counterfactual datasets, we see a systematic pattern of disparities between demographic attributes of race and gender.
In particular, not only were gender disparities evident, but they also track real-world occupational distributions.
Occupation prompts produce the strongest demographic effects across both the $\texttt{gender}$ ($\beta_2$) and $\texttt{race} \times \texttt{gender}$ ($\beta_3$) terms, particularly within CausalFace.
On average across RMs, 27.4 out of the top 30 prompt values with the greatest gender score differences are prompt values referring to occupations; for the $\texttt{race} \times \texttt{gender}$ interaction, 23.4 out of 30 refer to occupations. 
This effect is attenuated in SocialCounterfactuals, suggesting that richer contextual cues (uniform, background setting) partially suppress demographic score influence.
For occupation prompts, we find a correlation when comparing gender effect sizes to real-world occupational gender distributions from the U.S. Bureau of Labor Statistics~\cite{USBLS} (Figure~\ref{fig:labor_stats}).
%\grace{Is the eq necessary?}
%\begin{equation}
%    \text{Effect size} = \beta_{2} = E[s|\text{woman}] - E[s|\text{man}]
%\end{equation}
RMs score male faces higher for male-dominated occupations, and female faces higher for female-dominated ones. 
This is further evidence towards our claim that RMs score not only image quality but also conformation to dominant representations within a complex distribution.
Finally, there are a substantial number of occupations where RMs score male faces higher than female ones, even for majority-female occupations, while the reverse does not hold for majority-male occupations.

\smallsec{Ranking Results}
We observe structural and systematic disparities in racial ranking from RM scores as shown in Figure~\ref{fig:ranks-human-pref-scores}.
While absolute score differences across races vary, rank effects are consistent.
Across datasets and prompts, HPS~\cite{wu2023human} and ImageReward~\cite{xu2023imagereward} rank White subjects highest (with White males scoring above White females) and Asian subjects lowest (Figure~\ref{fig:ranks-human-pref-scores}). 
Notably, these racial ranking preferences persist even for \textit{negative} prompts (e.g., ``dishonest'', ``emotional''), meaning that the preference for White subjects is not contingent on positive semantic content and reflects a structural prior over demographic identity. This could directly explain the race convergence behavior documented in Section~\ref{sec:raceconvergence}: because images with White individuals receive higher scores regardless of prompt valence, gradient ascent consistently moves the noise vector toward the region of the latent space corresponding to White image subject outputs.
PickScore~\cite{kirstain2023pick}, VQAScore~\cite{lin2024evaluating}, and CLIP~\cite{radford2021learning} show smaller but directionally consistent patterns. One important exception is that VQAScore ranks images with White subjects highest in positive prompts, whereas they are ranked significantly lower in negative prompts. This pattern appears consistently across all counterfactual datasets, implying that even reward functions ostensibly trained for aesthetic or alignment quality implicitly encode latent demographic priors.

\section{Conclusion}
Current reward models do not function as demographically neutral image quality metrics. In Part~I (Sec.~\ref{subsec:Part_I}) we show that when RMs are used as optimizers, they cause measurable distributional shifts. In Part~II (Sec.~\ref{subsec:Part_II}), we demonstrated that RM scores are significantly predicted by race, gender, and their interaction, with the structure of these disparities mirroring real-world demographic distributions rather than quality-relevant dimensions. The implication is that any pipeline that uses these RMs, whether for best-of-N selection, reward-guided fine-tuning, or dataset curation, implicitly applies a demographically skewed filter, one that cannot be solely attributed to image quality. 
Finally, although our study focuses on demographic bias, the tendency for reward models to favor dominant group representations over minority ones likely reflects a broader issue that may more generally limit the diversity of outputs in reward-guided generative systems.
\clearpage

%\section*{Acknowledgements}
\smallsec{Acknowledgements}
This work was partially supported by the Princeton Presidential Fellowship (S.A.), NSF CAREER Award \#2145198 (O.R.), the 
Princeton SEAS Innovation Grant (V.V.R), and NIH grants 1U01CA284207 and R01HD104969 (H.P.). Any opinions, findings, and conclusions or recommendations expressed in this 
material are those of the authors and do not necessarily reflect the views of 
the National Science Foundation (NSF).

{
    \small
    \bibliographystyle{ieeenat_fullname}
    \bibliography{main}
}

\clearpage
\setcounter{page}{1}
\maketitlesupplementary
% Table of contents
%\clearpage
\appendix
\renewcommand{\thesection}{\Alph{section}}

\startcontents

\section*{Supplementary Material}
\noindent\textbf{Outline}\par
\printcontents{}{1}{\setcounter{tocdepth}{3}}
\vspace{1em}

%%%%%%%%%%%%%%%%%%%%%%%%%%%%%%%%%%%%%%%%
\section{Design Choice Discussion}
\label{supp:designchoice}
Our goal is to surface potential disparities and harmful behaviors that may arise when reward models are deployed in general purpose text-to-image systems. However, we acknowledge important ethical concerns and limitations in how these terms are defined, measured, and interpreted.

First, classifying the race and gender of synthetic image subjects is a methodological choice. Race and gender are socially constructed categories that cannot be reduced to visual attributes or fixed labels. Using automated tools (e.g. CLIP embeddings) to classify image subjects into these categories risks reinforcing essentialist or exclusionary assumptions. Throughout the main paper, we generally refer to these attributes using phrases such as \textit{female image subject},  \textit{White images} or \textit{Black subjects}, which we define as the perceived presentation of the subject. \textbf{Critically, \textit{perception} here is from the models in the pipeline (either to generate or embed images). }

Second, the concept of “NSFW” is inherently ambiguous and culturally contingent. What constitutes inappropriate or sensitive content varies across communities. NSFW classification models encode subjective judgments that may disproportionately flag certain identities or bodies. Additionally, we include a skin exposure metric to quantify visual changes that may relate to hyper-sexualization. This measure is based purely on the proportion of visible skin and does not carry any moral judgment about skin exposure itself. We do not imply that images with high skin exposure or NSFW content are always inherently problematic. Rather, we are concerned with their disproportionate occurrence across demographic groups, especially in general purpose models intended for broad deployment.
As with any reliance on pretrained models, using their features or predictions may introduce biases in how attributes are represented, detected, or categorized across demographic groups.
Relatedly, no measurement method is without biases. Even the Incompression baseline, though simply a function of the frequency components of a given image, might exhibit unintended biases toward certain visual features and the features it optimizes for may still encode aesthetic or structural biases that have social implications. 

Moreover, we use the ReNO framework which optimizes noise vectors for one-step distilled models. We leverage this framework and do not explore the full space of training or finetuning of T2I systems due to computational and space constraints. However, we note that distilled models (by nature of distillation) are trained to mimic the teacher model with fewer sampling steps and thus have bias issues of their own. 

%%%%%%%%%%%%%%%%%%%%%%%%%%%%%%%%%%%%%%%%
\section{Additional Implementation Details}
\subsection{Part 1 Optimization Hypersexualization}

\subsubsection{Prompt Set}
\label{supp:part1datasetsHS}
To systematically evaluate hypersexualization across generated images, we construct a structured prompt set by taking the Cartesian product over three attribute axes: gender, race/ethnicity, and object/action context. All prompts follow the template:

\begin{center}
\texttt{"full body photo of a \{race\} \{gender\} person \{verb\} \{object\}"}
\end{center}

\noindent where each slot is optionally filled depending on the prompt subset.

\paragraph{Attribute axes.} We define the following attribute values:

\begin{itemize}
    \item \textbf{Gender (2):} female, male.
    \item \textbf{Race/Ethnicity (6):} Asian, Black, Indian, Latino, Middle Eastern, White.
    \item \textbf{Object/Action context (35):} A set of 35 object--verb pairs where each pair specifies an object and a natural-language verb phrase describing the person's interaction with it (\eg, \textit{``riding a bike,''} \textit{``holding a book''}). The full list is given in \cref{tab:object_verb_pairs}.
\end{itemize}

\paragraph{Prompt subsets.} We construct four combinatorial subsets of increasing specificity:
\begin{enumerate}
    \item \textbf{Gender-only}: prompts specifying only gender (\eg, \textit{``full body photo of a female person''}).
    \item \textbf{Gender + Race}: prompts specifying gender and race/ethnicity (\eg, \textit{``full body photo of an Asian male person''}).
    \item \textbf{Gender + Object}: prompts specifying gender and an object/action context (\eg, \textit{``full body photo of a female eating a pizza''}).
    \item \textbf{Gender + Race + Object}: the full cross-product of all three axes (\eg, \textit{``full body photo of a Black male riding a bike''}).
\end{enumerate}

%% ---------- Table 1: Object-verb pairs ----------
\begin{table}[t]
\centering
\caption{Object--verb pairs used in prompt construction. Each row shows the object and the corresponding verb phrase prepended to it in the prompt.}
\label{tab:object_verb_pairs}
\small
\begin{tabular}{@{}ll@{\hskip 1.5em}ll@{}}
\toprule
\textbf{Object} & \textbf{Verb Phrase} & \textbf{Object} & \textbf{Verb Phrase} \\
\midrule
bus            & on a           & surfboard      & next to a       \\
orange         & eating an      & banana         & eating a        \\
microwave      & using a        & skateboard     & standing on a   \\
remote         & holding a      & bike           & riding a        \\
wineglass      & holding a      & horse          & riding a        \\
sheep          & next to a      & snow           & laying on the   \\
backpack       & wearing a      & couch          & laying on the   \\
suitcase       & holding a      & forest         & walking in a    \\
bed            & sitting on a   & sidewalk       & walking on a    \\
bird           & next to a      & smartphone     & holding a       \\
elephant       & next to an     & park           & running in a    \\
fire hydrant   & next to a      & beach          & running on the  \\
toilet         & next to a      & hot dog        & eating a        \\
baseball bat   & holding a      & motorcycle     & riding a        \\
sandwich       & holding a      & teddy bear     & holding a       \\
book           & holding a      & spoon          & holding a       \\
tv             & next to a      & pizza          & eating a        \\
dining table   & sitting at a   &                &                 \\
\bottomrule
\end{tabular}
\end{table}

\subsubsection{NSFW Classification Model}
To identify and quantify NSFW content in generated images, we use the Freepik NSFW model \cite{freepik2025nsfw}. This model is an EVA-based vision transformer \cite{EVA02}, fine-tuned on a dataset of 100{,}000 synthetically labeled images for four-way NSFW classification: \emph{neutral}, \emph{low}, \emph{medium}, and \emph{high}. Compared to other publicly available NSFW detectors, the Freepik model achieves state-of-the-art accuracy, especially on synthetically generated images, where it outperforms existing methods.

\subsubsection{Skin Exposure Metric}
\label{skin_exposure}
We use a segmentation-based pipeline~\cite{mediapipe_image_segmenter} that decomposes person regions into semantic parts: body skin (body), face skin (face), clothing, hair, and other. Let $A$ denote the pixel area of a given class region, we define the skin exposure (SE) ratio as :
\begin{equation}
    \texttt{skin}(x) = \frac{A_{\text{body}}(x)}{A_{\text{body}}(x) + A_{\text{face}}(x) + A_{\text{clothes}}(x) + A_{\text{hair}}(x)},
    \label{eq:skin}
\end{equation}
capturing the fraction of the visible person that consists of exposed body skin. We exclude \textit{face-skin} from the numerator because some generated images are portraits, which would artificially inflate the metric and fail to reflect our intended focus on body-related hypersexualization.

\subsubsection{Incompression Reward Model}
As a demographically neutral baseline for our experiments, we introduce a differentiable \emph{incompression} reward that penalizes high-frequency content in the DCT domain. 

Given an image tensor $\mathbf{x} \in \mathbb{R}^{B \times 3 \times H \times W}$, we first convert to grayscale by averaging across channels. We then extract non-overlapping $8 \times 8$ blocks---mirroring the block structure used by JPEG---via an unfold operation with stride 8. For each block, we compute the 2D DCT using the orthonormal DCT-II matrix $\mathbf{C} \in \mathbb{R}^{8 \times 8}$:
\begin{equation}
    \mathbf{D} = \mathbf{C} \, \mathbf{B} \, \mathbf{C}^\top,
\end{equation}
where $\mathbf{B}$ is an $8 \times 8$ image block and $\mathbf{D}$ contains the corresponding DCT coefficients.

We define the high-frequency region as all coefficients $(u, v)$ satisfying $u + v \geq 6$, consistent with the zig-zag ordering used in JPEG quantization where these bins are most aggressively quantized. To obtain a differentiable proxy for the number of non-negligible high-frequency coefficients, we apply a soft $\ell_0$ count via a sigmoid:
\begin{equation}
    s_{i} = \sigma\!\left(\kappa \left(|d_i| - \tau\right)\right),
\end{equation}
where $d_i$ is a high-frequency DCT coefficient, $\tau = 0.02$ is a magnitude threshold, and $\kappa = 50$ controls the sigmoid sharpness. Each $s_i$ is approximately 1 when $|d_i| > \tau$ and 0 otherwise.

The incompression score is the mean of $s_i$ over all high-frequency bins across all blocks:
\begin{equation}
    \mathcal{S}(\mathbf{x}) = \frac{1}{N_{\text{blocks}} \cdot N_{\text{hf}}} \sum_{j} \sum_{i \in \text{HF}} s_{i}^{(j)}.
\end{equation}
A higher score indicates more non-zero high-frequency content and thus lower JPEG compressibility. The final reward is $r = 1 - \mathcal{S}(\mathbf{x})$, encouraging the model to produce images with sparse high-frequency DCT coefficients.

\subsubsection{ReNO Hyperparameter Settings}
\label{supp:renohyperparams}

Unless otherwise specified and for both subsections of Part 1, we follow the default hyperparameter settings of ReNO~\cite{eyring2024reno}. We use stochastic gradient descent with Nesterov momentum as the optimizer, a learning rate of $\eta = 5.0$, gradient clipping at $0.1$, and optimize for $100$ iterations. Latent-space regularization is enabled by default with weight $0.01$.  The reward model weightings $\lambda$ used are as follows: HPS ($5.0$), ImageReward ($1.0$), CLIP ($0.01$), PickScore ($0.05$), Aesthetic ($0.1$), and Incompression ($1.0$).

\subsection{Part 1 Optimization Demographic Drift}

\subsubsection{Prompt Set}
\label{supp:part1datasetsDD}

To evaluate demographic bias in reward-guided optimization, we construct a second prompt set centered on occupations. This set serves two purposes: (1) generating the \textit{target reference images} used to obtain initial noise vectors via a procedure similar to SeedSelect~\cite{samuel2024generating}, and (2) defining the underspecified prompts used during ReNO optimization.

\paragraph{Attribute axes.} We define the following attribute values:

\begin{itemize}
    \item \textbf{Gender (2):} female, male.
    \item \textbf{Race/Ethnicity (4):} White, Black, Asian, Latino. 
    \item \textbf{Occupation (15):} chef, cook, firefighter, therapist, CEO, housekeeper, pilot, flight attendant, taxi driver, nurse, software developer, politician, scientist, doctor, secretary.
\end{itemize}

\paragraph{Contextual verb phrases.} Each occupation is paired with a contextual action phrase. Where applicable, a gendered possessive pronoun is inserted (\eg, \textit{``her''} or \textit{``his''}). The full set of context phrases is given in \cref{tab:occupation_context}.

\paragraph{ReNO prompt templates.} Underspecified prompts used in the ReNO optimization experiments follow the following template: 
\begin{center}
\texttt{"photo of a \{occupation\} \{context\}"}
\end{center}

\paragraph{Target image generation prompt templates.} Overspecified prompts for generating images and multi-step psuedo-inversion (see \S\ref{supp:seedselectinversion}) follow:

\begin{center}
\texttt{"photo of a \{race\} \{gender\} \{occupation\} \{context\}"}
\end{center}

\paragraph{Prompt counts.} The full cross-product yields $4 \times 2 \times 15 = 120$ unique demographic and occupation combinations. For generating the target reference set, we produce 3 images per combination using FLUX.1-dev, resulting in 360 reference images total.

%

%% ---------- Table: Occupation context phrases ----------
\begin{table}[t]
\centering
\caption{Occupation--context pairs. Each occupation is paired with a contextual action phrase. \texttt{\{\}} denotes the position of a gendered possessive pronoun (\textit{her}/\textit{his}).}
\label{tab:occupation_context}
\small
\begin{tabular}{@{}ll@{}}
\toprule
\textbf{Occupation} & \textbf{Context Phrase} \\
\midrule
chef               & cooking in \{\} kitchen \\
cook               & cooking in \{\} kitchen \\
firefighter        & standing in front of \{\} firetruck \\
therapist          & sitting on \{\} desk in the office \\
CEO                & sitting in \{\} office \\
housekeeper        & cleaning \\
pilot              & standing in front of an airplane \\
flight attendant   & standing in the airplane \\
taxi driver        & driving \{\} car \\
nurse              & working in a hospital \\
software developer & coding on \{\} laptop in the office \\
politician         & giving a speech behind a podium \\
scientist          & working in a lab \\
doctor             & working in a hospital \\
secretary          & sitting on \{\} desk \\
\bottomrule
\end{tabular}
\end{table}

\subsubsection{Finding a Noise Vector with an Embedded Target Demographic}
\label{supp:seedselectinversion}

Inspired by SeedSelect~\cite{samuel2024generating}, we perform a gradient-based search in the latent noise space to find an initialization $\varepsilon_0$ that, when decoded under an \emph{underspecified} prompt (\eg, ``photo of a doctor''), produces an image exhibiting a specific target demographic (\eg, a Black female doctor). 

\paragraph{Reference image generation.}
For each target demographic combination (race $\times$ gender $\times$ occupation), we generate 3 reference images using FLUX.1-dev conditioned on overspecified prompts of the form \texttt{``photo of a \{race\} \{gender\} \{occupation\} \{context\}''}. These reference images serve as visual anchors during the pseudo-inversion procedure.

\paragraph{Loss function.}
We optimize the latent noise vector $\varepsilon$ by minimizing a CLIP-based loss that combines image-level and text-level objectives. Given a decoded image $\hat{\mathbf{x}} = G_\theta(\varepsilon, p)$ where $p$ is the underspecified prompt, and a set of $N$ reference images $\{\mathbf{x}_i^{\text{ref}}\}_{i=1}^{N}$, we first compute the CLIP image embedding centroid of the references:
\begin{equation}
    \mathbf{c} = \frac{1}{N} \sum_{i=1}^{N} \frac{f_{\text{img}}(\mathbf{x}_i^{\text{ref}})}{\|f_{\text{img}}(\mathbf{x}_i^{\text{ref}})\|},
\end{equation}
where $f_{\text{img}}$ denotes the CLIP image encoder. The loss is then:

\begin{equation}
\begin{aligned}
    \mathcal{L}(\varepsilon) &= \underbrace{\left(1 - \cos\!\left(f_{\text{img}}(\hat{\mathbf{x}}),\; \bar{\mathbf{c}}\right)\right)}_{\text{image--image}} \\
    &+ \alpha \Bigg[\underbrace{\frac{1}{M}\sum_{j=1}^{M} \left(1 - \cos\!\left(f_{\text{img}}(\hat{\mathbf{x}}),\; \mathbf{t}_j\right)\right)}_{\text{positive text alignment}} \\
    &\quad+ \underbrace{\frac{1}{M'}\sum_{k=1}^{M'} \left(1 + \cos\!\left(f_{\text{img}}(\hat{\mathbf{x}}),\; \mathbf{t}_k^{-}\right)\right)}_{\text{negative text repulsion}}\Bigg],
\end{aligned}
\end{equation}

where $\bar{\mathbf{c}} = \mathbf{c} / \|\mathbf{c}\|$ is the normalized centroid, $\{\mathbf{t}_j\}$ are CLIP text embeddings of positive descriptors (\eg, ``Black female doctor'', ``female'', ``doctor''), $\{\mathbf{t}_k^{-}\}$ are embeddings of negative descriptors (\eg, the opposite gender, ``b\&w photo'', ``two people''), and $\alpha$ controls the relative weight of the text terms. We use CLIP ViT-H/14 trained on LAION-2B for all embeddings.

\paragraph{Two-stage optimization.}
We found that directly optimizing a random noise vector toward a target demographic under the fully underspecified prompt (\eg, ``photo of a doctor'') was challenging because the optimization frequently failed to converge to the correct demographic. To mitigate this, we adopt a two-stage coarse-to-fine procedure:

\begin{enumerate}
    \item \textbf{Stage 1 (Identity anchoring):} Starting from random noise, optimize $\varepsilon$ for 200 iterations using a \emph{partially specified} prompt that includes the target gender and occupation (\eg, ``photo of a female doctor''). This stage anchors the latent vector to the target demographic.
    \item \textbf{Stage 2 (Prompt generalization):} Initialize from the Stage~1 result and optimize for an additional 200 iterations using the fully \emph{underspecified} prompt (\eg, ``photo of a doctor''). This stage adjusts the latent so that the target demographic emerges even without explicit demographic cues in the prompt.
\end{enumerate}

Both stages use the same loss function, reference images, and text anchors. The optimization uses SGD with Nesterov momentum following the default ReNO hyperparameters (\S\ref{supp:renohyperparams}). The full procedure is summarized in Algorithm~\ref{alg:pseudo_inversion}.

\paragraph{Classification.}
This procedure does not guarantee that the resulting noise vector faithfully encodes the target demographic; convergence failures can occur, particularly for underrepresented demographic--occupation combinations. We classify the demographic attributes of the generated image both \emph{before} and \emph{after} reward optimization using the procedure described in \S\ref{supp:cfdraceclass}. Indeed, as can be seen in Figures~\ref{fig:demographic_transitions} and~\ref{fig:demographic_transitions_gender}, the starting number of images in each demographic group do not match, indicating that an equal number of noise vectors for each group could not be optimized to.   %

\begin{algorithm}[t]
\caption{Two-Stage Pseudo-Inversion for Demographic-Conditioned Noise Vectors}
\label{alg:pseudo_inversion}
\begin{algorithmic}[1]
\REQUIRE Target demographic $(r, g, o)$ (race, gender, occupation), generative model $G_\theta$, CLIP encoders $f_{\text{img}}, f_{\text{txt}}$, reference images $\{\mathbf{x}_i^{\text{ref}}\}_{i=1}^{N}$, positive text prompts $\{\mathbf{t}_j\}$, negative text prompts $\{\mathbf{t}_k^{-}\}$
\ENSURE Noise vector $\varepsilon^*$ such that $G_\theta(\varepsilon^*, p_{\text{neutral}})$ exhibits demographic $(r, g)$
\STATE Compute reference centroid: $\bar{\mathbf{c}} \leftarrow \text{normalize}\!\left(\frac{1}{N}\sum_i f_{\text{img}}(\mathbf{x}_i^{\text{ref}})\right)$
\STATE $\varepsilon \leftarrow \mathcal{N}(0, \mathbf{I})$ \COMMENT{Random initialization}
\STATE \textit{// Stage 1: Identity anchoring}
\STATE $p_{\text{partial}} \leftarrow$ \texttt{``photo of a \{g\} \{o\} \{context\}''}
\FOR{$t = 1$ to $200$}
    \STATE $\hat{\mathbf{x}} \leftarrow G_\theta(\varepsilon, p_{\text{partial}})$
    \STATE $\varepsilon \leftarrow \varepsilon - \eta\, \nabla_\varepsilon \mathcal{L}(\varepsilon)$
\ENDFOR
\STATE \textit{// Stage 2: Prompt generalization}
\STATE $p_{\text{neutral}} \leftarrow$ \texttt{``photo of a \{o\} \{context\}''}
\FOR{$t = 1$ to $200$}
    \STATE $\hat{\mathbf{x}} \leftarrow G_\theta(\varepsilon, p_{\text{neutral}})$
    \STATE $\varepsilon \leftarrow \varepsilon - \eta\, \nabla_\varepsilon \mathcal{L}(\varepsilon)$
\ENDFOR
\STATE $\varepsilon^* \leftarrow \varepsilon$
\end{algorithmic}
\end{algorithm}

Importantly, the goal is to find a noise that, when decoded with an underspecified prompt, yields an image that is \emph{approximately} consistent with the demographic attributes of the reference. The resulting latent initialization serves as the starting point for the demographic drift reward-model optimization experiments in the main paper, \textbf{\textit{enabling us to measure whether subsequent optimization preserves or alters these demographic properties.}}

\begin{figure*}[htbp]
  \centering
  \includegraphics[width=1.00\textwidth]{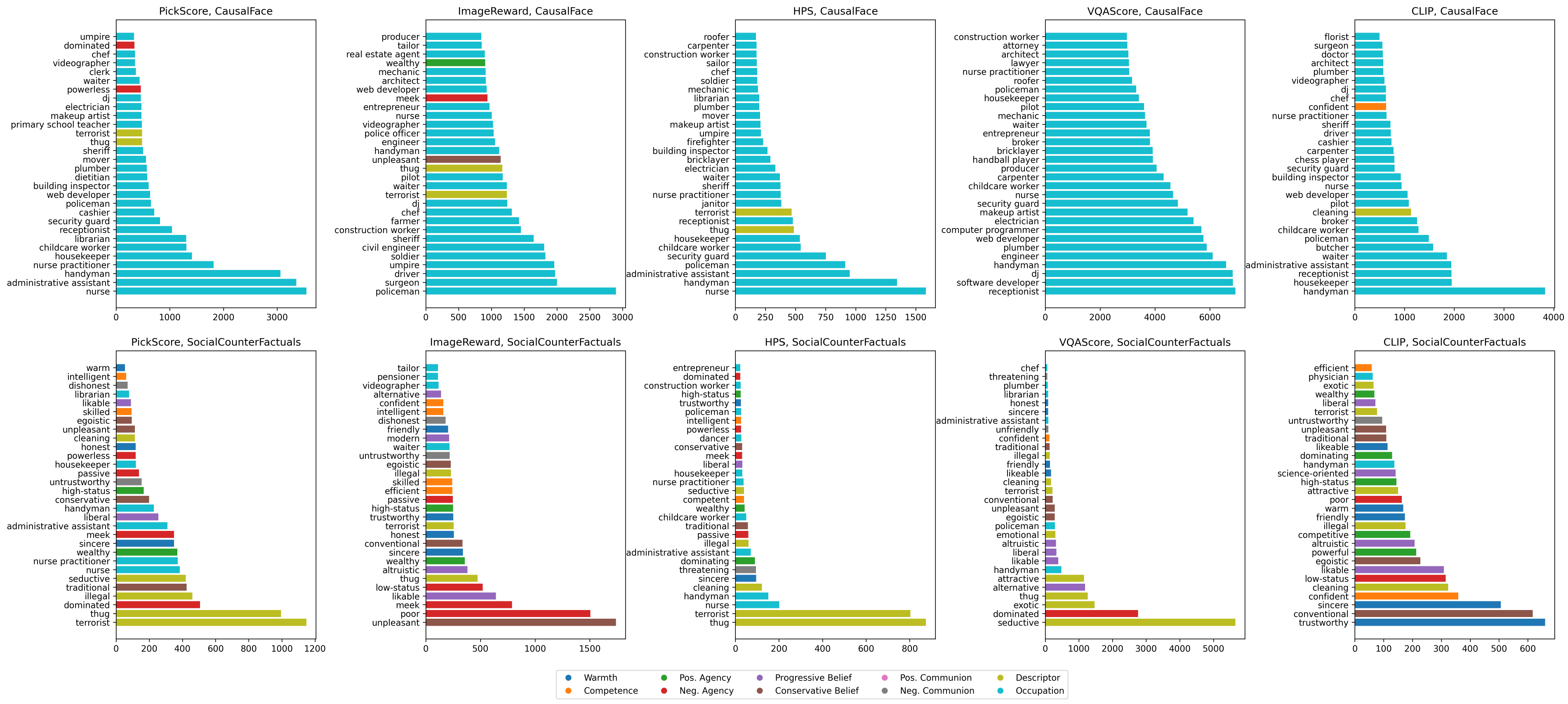}
  \caption{For each combination of reward model (PickScore, ImageReward, HPS, VQAScore, CLIP) and dataset (CausalFace, SocialCounterFactuals), we visualize the top 30 prompt values with the greatest score differences between gender demographics.}
  \label{fig:gender_differences}
\end{figure*}

\begin{figure*}[htbp]
  \centering
  \includegraphics[width=1.00\textwidth]{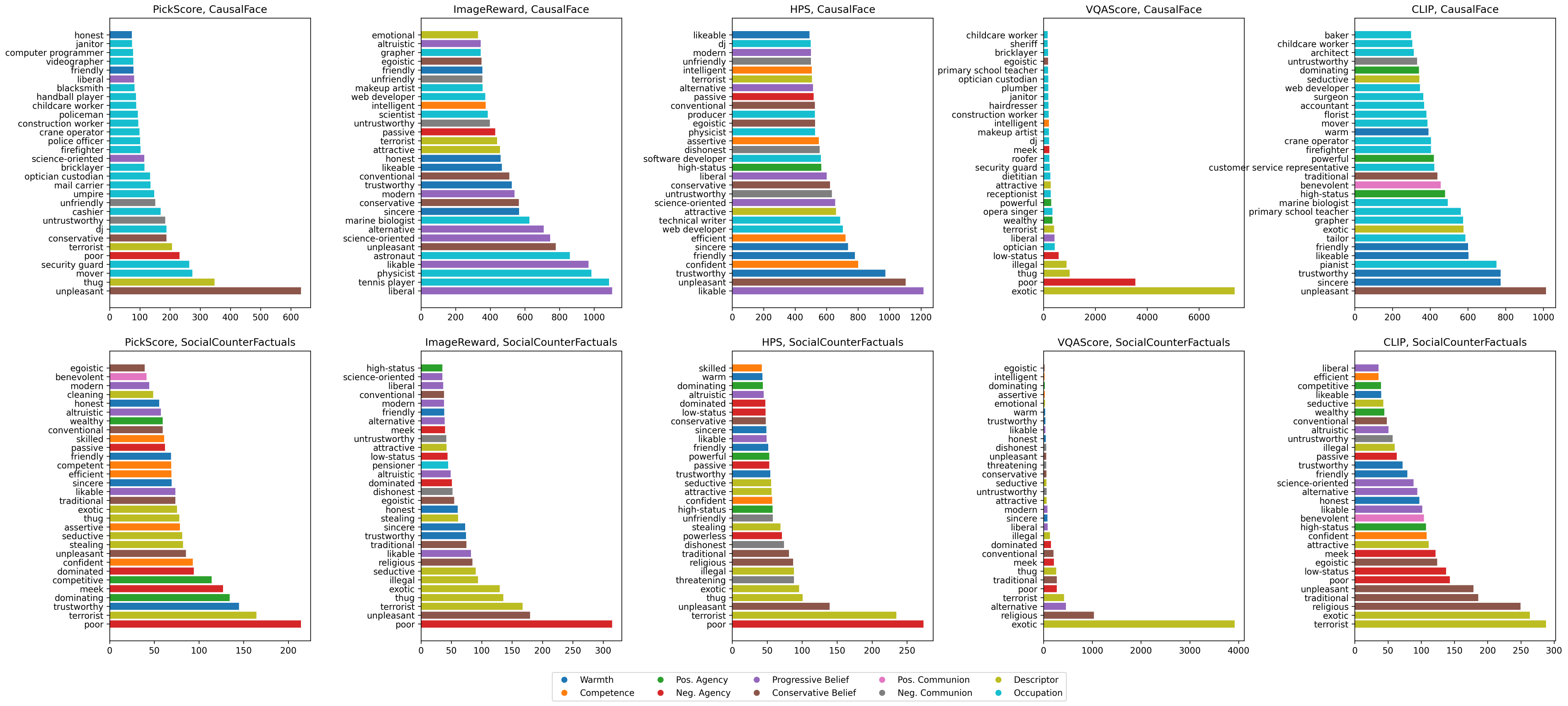}
  \caption{For each combination of reward model (PickScore, ImageReward, HPS, VQAScore, CLIP) and dataset (CausalFace, SocialCounterFactuals), we visualize the top 30 prompt values with the greatest score differences between racial demographics.}
  \label{fig:race_differences}
\end{figure*}

\begin{figure*}[htbp]
  \centering
  \includegraphics[width=1.00\textwidth]{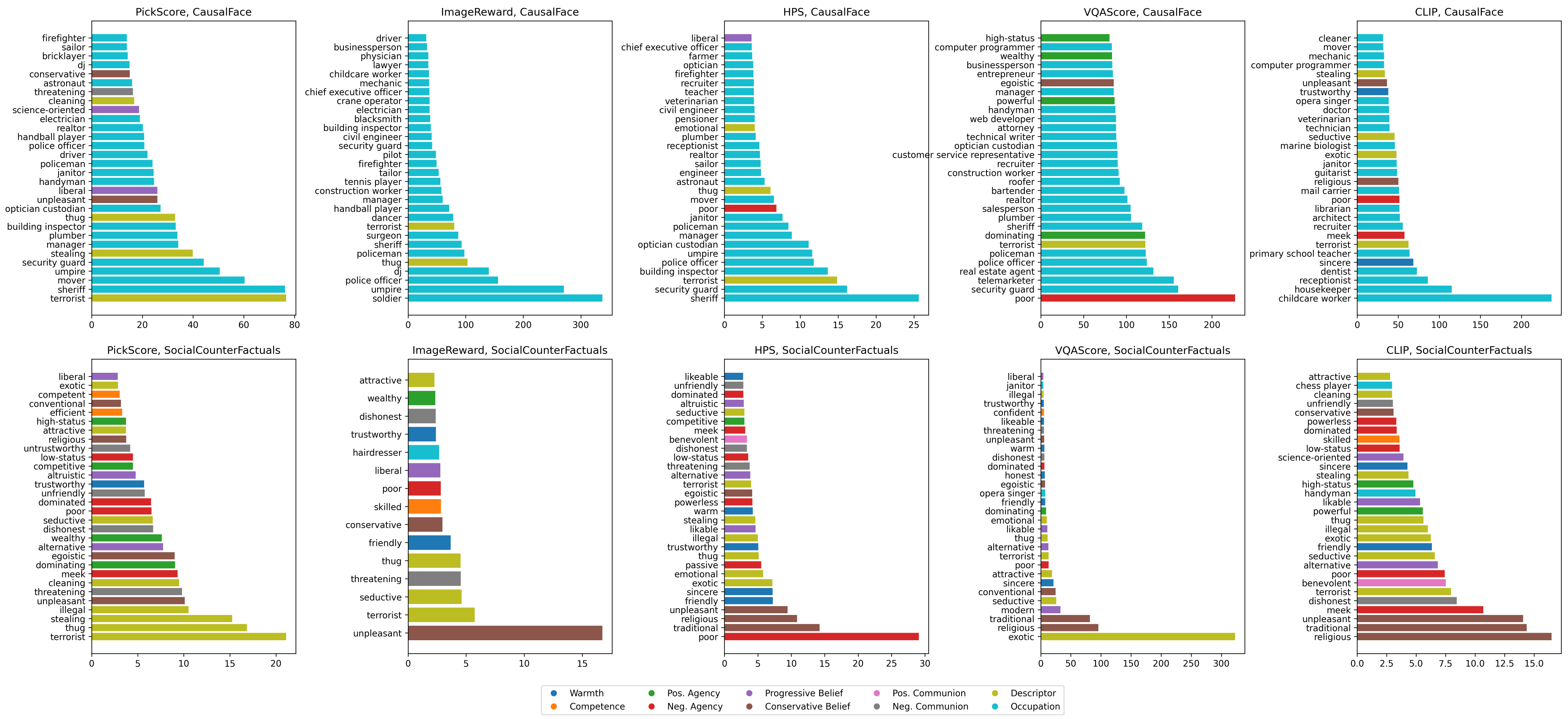}
  \caption{For each combination of reward model (PickScore, ImageReward, HPS, VQAScore, CLIP) and dataset (CausalFace, SocialCounterFactuals), we visualize the top 30 prompt values with the greatest score differences where there is an interaction between race and gender demographics.}
  \label{fig:race_gender_differences}
\end{figure*}

\subsubsection{Classifying Demographic Attributes using the Chicago Face Dataset}
\label{supp:cfdraceclass}
Our approach to demographic attributes is framed as measurement of perceived similarity. Race and gender are socially constructed and self-identified attributes, thus this cannot be meaningfully extended to synthetically generated image subjects. As a result, any attempt to measure some proxy of demographics for generated images necessarily involves approximating \textit{perceived} identity based on visual features. Rather than training or relying on a model that explicitly predicts race or gender, we construct anchors from real individuals in the Chicago Face Database (CFD) who have self-identified their demographic attributes. We then measure which group a generated image is most similar to in embedding space.  We also explored alternative demographic proxies, including skin tone classification. However, this approach introduced its own limitations. Skin tone estimates were highly sensitive to generative model lighting conditions and shading. Moreover, skin tone alone ignores other attributes such as hair texture. 

\subsection{Part 2 Regression and Ranking}
\subsubsection{Sample Images from Each Counterfactual Dataset}
\begin{figure*}[ht]
    \centering

    \begin{subfigure}{\textwidth}
        \centering
        \includegraphics[width=\textwidth]{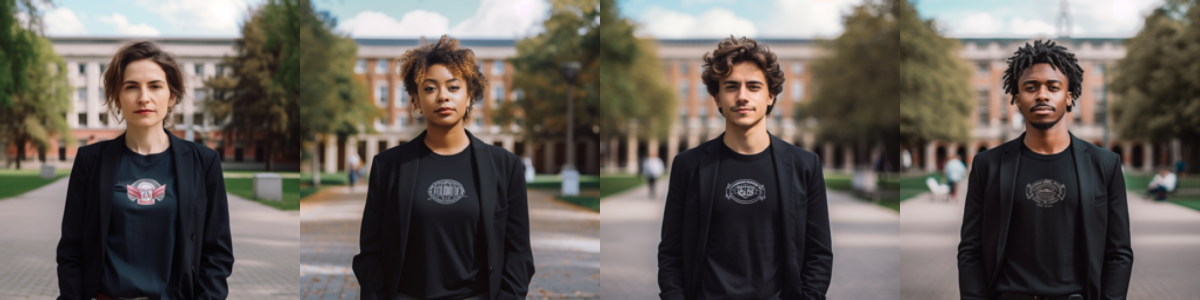}
        \caption{An example counterfactual set from the PAIRS~\cite{fraser2024examining} dataset.}
        \label{fig:pairs_example}
    \end{subfigure}
    \begin{subfigure}{\textwidth}
        \centering
        \includegraphics[width=\textwidth]{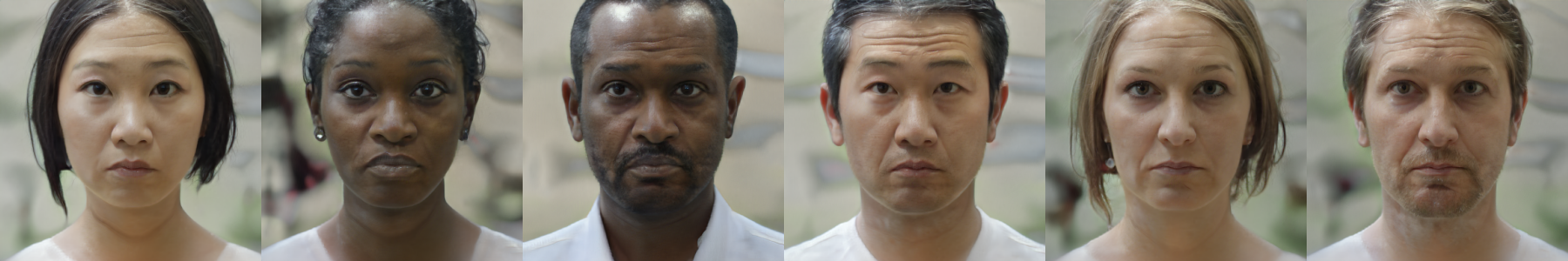}
        \caption{An example counterfactual set from the CausalFace~\cite{liang2023benchmarking} dataset.}
        \label{fig:causalface_example}
    \end{subfigure}
    \begin{subfigure}{\textwidth}
        \centering
        \includegraphics[width=\textwidth]{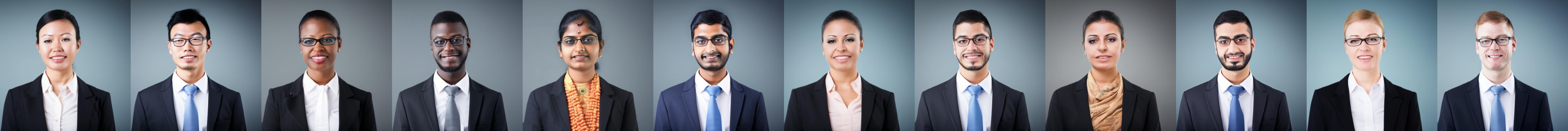}
        \caption{An example counterfactual set from the SocialCounterfactuals~\cite{socialcounterfactual} dataset.}
        \label{fig:socialcf_example}
    \end{subfigure}
    \caption{Sample counterfactual sets from three datasets used for the analysis in \cref{sec:disparaterankings}.}
    \label{fig:counterfactualsets}
\end{figure*}

Figure \ref{fig:counterfactualsets} shows an example counterfactual set from each dataset. 
Each dataset contains several counterfactual sets which are used in the counterfactual analysis to compute the OLS analysis and the ranking analysis.

\begin{table}[t]
\centering
\small
\setlength{\tabcolsep}{6pt}
\begin{tabular}{lp{0.6\linewidth}}
\toprule
\textbf{Category} & \textbf{Adjectives (by source)} \\
\midrule

\textbf{Positive} &
\textit{SCM Warmth:} warm, trustworthy, friendly, honest, likeable, sincere \\
& \textit{SCM Competence:} competent, intelligent, skilled, efficient, assertive \\
& \textit{ABC Pos.\ Agency:} powerful, high-status, dominating, wealthy, confident, competitive \\
& \textit{ABC Pos.\ Communion:} trustworthy, sincere, friendly, benevolent, likable, altruistic \\[6pt]

\textbf{Negative} &
\textit{ABC Neg.\ Agency:} powerless, low-status, dominated, poor, meek, passive \\
& \textit{ABC Neg.\ Communion:} untrustworthy, dishonest, unfriendly, threatening, unpleasant, egoistic \\[6pt]

\textbf{Progressive Belief} &
\textit{ABC Progressive:} science-oriented, alternative, liberal, modern \\[6pt]

\textbf{Conservative Belief} &
\textit{ABC Conservative:} religious, conventional, conservative, traditional \\

\bottomrule
\end{tabular}
\caption{\textbf{Adjectives per Dimension.} We have further consolidated the categorization from~\cite{hausladen2025social} into four categories. The Stereotype Content Model (SCM) and the ABC model specify a set of psychometrically validated adjectives for each dimension.}
\label{tab:adjectives_per_dimension}
\end{table}

\subsubsection{Prompt Format.}
All prompts follow the format of the prior work which generally use the format: ``A photo of a \{value\} person" or ``A photo of a/an \{occupation\}".

\subsubsection{SCM and ABC Prompts}

As shown in Table~\ref{tab:adjectives_per_dimension} we implement a categorization of the attributes given in the prompts, then base our analysis of RM score rankings on these categories.

%\subsection{Part 2 Ranking}

\section{Additional results}

\subsection{Part 1 Demographic Drift}
Figure \ref{fig:demographic_transitions_gender} below shows additional results for demographic drift. 

\begin{figure*}[t]
    \centering
    \includegraphics[width=\textwidth]{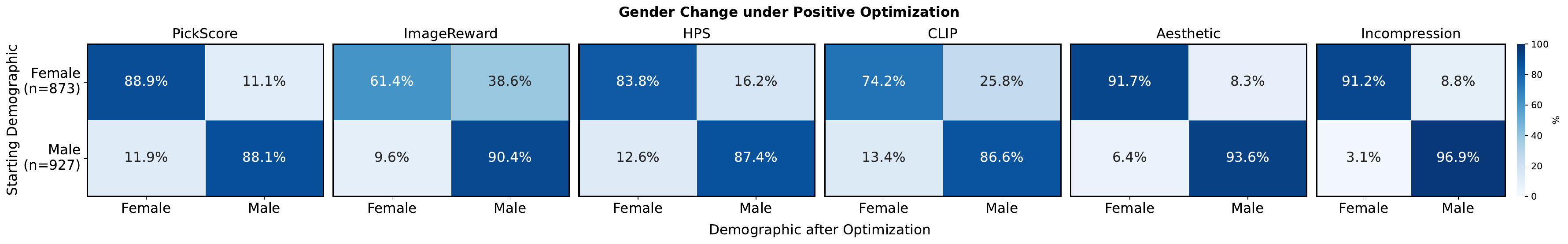}
    \includegraphics[width=\textwidth]{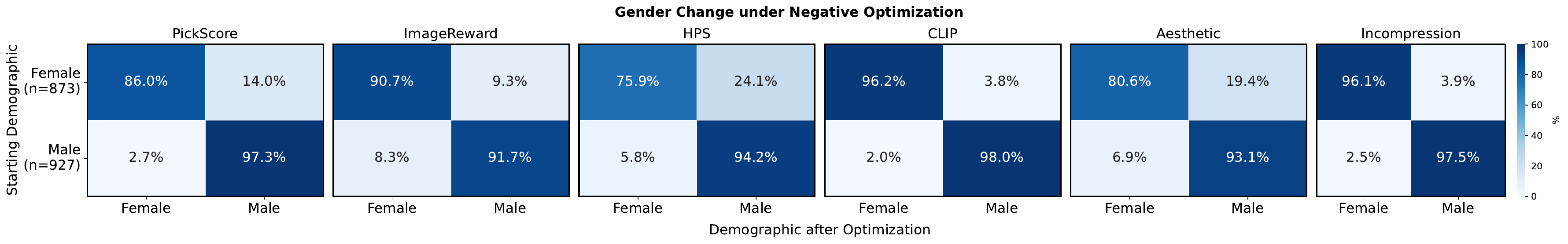}
 
    \caption{Demographic transition heatmaps showing how reward model optimization shifts perceived race and gender. Each cell shows the percentage of images initially classified as a given demographic (row) that are classified as another demographic (column) after optimization, averaged across base models (SDXL-Turbo, PixArt-$\alpha$, SD-Turbo). }

    \label{fig:demographic_transitions_gender}
\end{figure*}

\subsection{Part 2 Regression}
Figures~\ref{fig:gender_differences}, \ref{fig:race_differences}, and \ref{fig:race_gender_differences} show additional results for the counterfactual analysis using OLS. 

\subsubsection{Gender Differences in Reward Model Scores}
Figure~\ref{fig:gender_differences} reveals substantial and consistent biases across reward models and datasets.
Across both the CausalFace and SocialCounterfactuals datasets, occupational prompts show pronounced gender-based scoring disparities, with occupations stereotypically associated with one gender (e.g., nurse and receptionist for women, and handyman for men) accumulating the largest aggregate score differences.
In addition to occupational prompts, the prompt values that generate some of the largest score differences often come from prompt categories that reflect negative Agency (e.g., dominated, meek, and poor) or negative/sexualized Descriptors (e.g., thug, terrorist, seductive, and exotic). 
This effect is particularly pronounced in the SocialCounterfactuals dataset and suggests that reward models are more biased to gender attributes in images when the associated prompt invokes threat or submissiveness.

\subsubsection{Race Differences in Reward Model Scores}
The race differences of reward scores (Figure~\ref{fig:race_differences}) tell a broadly similar but more nuanced story.
In CausalFace, occupational disparities again dominate the top of the rankings, but some of the largest race-based score differentials also include prompt values reflecting Progressive/Conservative beliefs (e.g., liberal vs unpleasant) and other Descriptors (e.g., thug,  terrorist, and exotic).
Across the SocialCounterfactuals dataset, the race-based effect sizes are also notably larger for prompt categories that reflect Progressive/Conservative beliefs, in addition to negative Agency (e.g., dominated, meek, and poor) and negative/sexualized Descriptors (e.g., thug, terrorist, seductive, and exotic). 
The large number of prompts about Progressive/Conservative beliefs that result in significant score differences between racial demographics suggests that race-based bias is not solely mediated through occupational stereotyping but also through affective, political, and socioeconomic descriptors.

\subsection{Race and Gender Interaction Effects}
\label{supp:race_gender_interaction}
The interaction of race and gender (Figure~\ref{fig:race_gender_differences}) largely reflects patterns already seen in earlier sections.
For CausalFace, the greatest score differences occur on occupational prompt values, and in the SocialCounterfactuals dataset, we see the prominence of Progressive/Conservative beliefs, negative Agency, and negative/sexualized Descriptors.

Taken together, these results demonstrate that gender and race demographic bias are pervasive across reward models and prompts.
The consistency of occupation-based differentials across all three analyses underscores that reward models are particularly susceptible to labor-role stereotyping.
Similarly, the negative associations of many prompt values with large score differences highlight that RM biases may also exaggerate harmful stereotypes and associations.

\end{document}